\documentclass[conference]{IEEEtran}
\usepackage{times}

\usepackage[numbers]{natbib}
\usepackage{multicol}
\usepackage[bookmarks=true]{hyperref}
\usepackage{tabularx}
\usepackage{array} % Add the 'array' package for column type definitions
\usepackage{graphicx}
\usepackage{subcaption}
\usepackage{multirow}
\usepackage{listings}
\usepackage{xcolor}
\usepackage{booktabs} 
\usepackage{comment}

\begin{document}

% paper title
\title{LHManip: A Dataset for Long-Horizon Language-Grounded Manipulation Tasks in Cluttered Tabletop Environments}

% You will get a Paper-ID when submitting a pdf file to the conference system
%\author{Author Names Omitted for Anonymous Review. Paper-ID [add your ID here]}
\author{\authorblockN{Federico Ceola$^{1}$, Lorenzo Natale$^{1}$, Niko S{\"u}nderhauf$^{2}$ and Krishan Rana$^{2}$}

%\thanks{$^{1}$Humanoid Sensing and Perception (HSP), Istituto Italiano di Tecnologia (IIT), Genoa, Italy.}
%\thanks{$^{2}$Queensland University of Technology Centre for Robotics (QCR), Brisbane, Australia.}
%}

%\author{\authorblockN{Federico Ceola, Lorenzo Natale}
\authorblockA{$^{1}$Humanoid Sensing and Perception (HSP), Istituto Italiano di Tecnologia (IIT), Genoa, Italy. \\ $^{2}$Queensland University of Technology Centre for Robotics (QCR), Brisbane, Australia.}}

%\author{\authorblockN{Michael Shell}
%\authorblockA{School of Electrical and\\Computer Engineering\\
%Georgia Institute of Technology\\
%Atlanta, Georgia 30332--0250\\
%Email: mshell@ece.gatech.edu}
%\and
%\authorblockN{Homer Simpson}
%\authorblockA{Twentieth Century Fox\\
%Springfield, USA\\
%Email: homer@thesimpsons.com}
%\and
%\authorblockN{James Kirk\\ and Montgomery Scott}
%\authorblockA{Starfleet Academy\\
%San Francisco, California 96678-2391\\
%Telephone: (800) 555--1212\\
%Fax: (888) 555--1212}}

% avoiding spaces at the end of the author lines is not a problem with
% conference papers because we don't use \thanks or \IEEEmembership

% for over three affiliations, or if they all won't fit within the width
% of the page, use this alternative format:
% 
%\author{\authorblockN{Michael Shell\authorrefmark{1},
%Homer Simpson\authorrefmark{2},
%James Kirk\authorrefmark{3}, 
%Montgomery Scott\authorrefmark{3} and
%Eldon Tyrell\authorrefmark{4}}
%\authorblockA{\authorrefmark{1}School of Electrical and Computer Engineering\\
%Georgia Institute of Technology,
%Atlanta, Georgia 30332--0250\\ Email: mshell@ece.gatech.edu}
%\authorblockA{\authorrefmark{2}Twentieth Century Fox, Springfield, USA\\
%Email: homer@thesimpsons.com}
%\authorblockA{\authorrefmark{3}Starfleet Academy, San Francisco, California 96678-2391\\
%Telephone: (800) 555--1212, Fax: (888) 555--1212}
%\authorblockA{\authorrefmark{4}Tyrell Inc., 123 Replicant Street, Los Angeles, California 90210--4321}}

\maketitle

\begin{abstract}
Instructing a robot to complete an everyday task within our homes has been a long-standing challenge for robotics. While recent progress in language-conditioned imitation learning and offline reinforcement learning has demonstrated impressive performance across a wide range of tasks, they are typically limited to short-horizon tasks -- not reflective of those a home robot would be expected to complete. While existing architectures have the potential to learn these desired behaviours, the lack of the necessary long-horizon, multi-step datasets for real robotic systems poses a significant challenge. To this end, we present the \textit{Long-Horizon Manipulation} (\textit{LHManip}) dataset comprising 200 episodes, demonstrating 20 different manipulation tasks via real robot teleoperation. The tasks entail multiple sub-tasks, including grasping, pushing, stacking and throwing objects in highly cluttered environments. Each task is paired with a natural language instruction and multi-camera viewpoints for point-cloud or NeRF reconstruction. In total, the dataset comprises 176,278 observation-action pairs which form part of the Open X-Embodiment dataset. The full \textit{LHManip} dataset is made publicly available \href{https://github.com/fedeceola/LHManip}{here}.
\end{abstract}

\IEEEpeerreviewmaketitle

%\begin{comment}

\section{Introduction}
Solving long-horizon manipulation tasks is crucial for addressing real-world applicability of robotic problems. Many practical tasks and activities, such as meal preparation, room cleaning, or workspace organization, involve a sequence of actions performed over an extended period. These tasks are inherently more complex than the short-term ones, as they require robots not only to manipulate objects but also to plan and execute actions across multiple steps. Long-horizon manipulation datasets have potential to allow the development of algorithms capable of generalizing across diverse scenarios, adapting to new settings, and addressing the challenges posed by tasks that require several steps to be executed.

\begin{figure}[t]
    \vspace{.6em}
    \centering
    \includegraphics[width=0.9\linewidth]{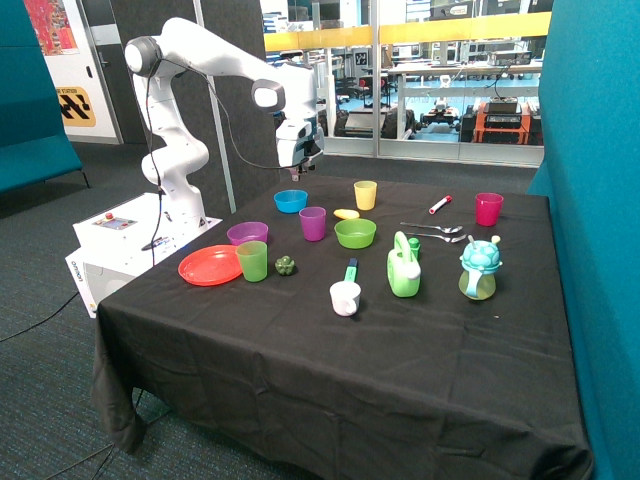}
    \caption{Robot and environment setup used for data collection.}
    \label{fig:lhmanip_im}
\end{figure}

Long-horizon manipulation tasks bring forth the importance of perceptual skills in robotic systems. While requiring to execute low-level control policies to solve the sub-tasks in which they are decomposed, these tasks also require high-level planning and reasoning capabilities. Furthermore, as robots become increasingly integrated into human-centered settings, they will be required to understand and follow natural language instructions for such tasks. Understanding natural language instructions offers the opportunity to develop robotic systems that are capable of autonomously learning, generalizing and adapting to novel settings and environments, rather than being explicitly programmed for each task~\cite{shridhar2023perceiver, pmlr-v229-zitkovich23a}.

\begin{figure*}[t!]
    \centering
    \includegraphics[width=0.8\linewidth]{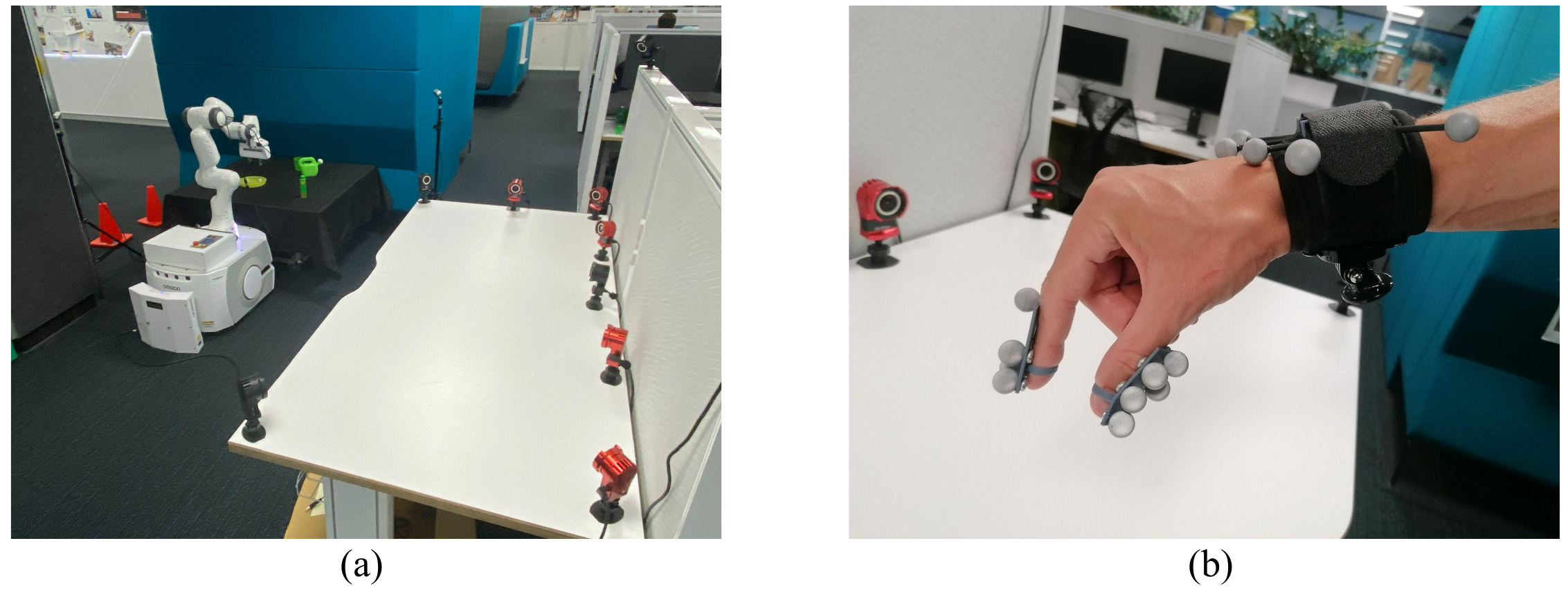}
    \caption{(a) Motion capture and robot setup. (b) The robot was teleoperated by a human operator equipped with a motion capture system for hand gestures and movements detection in the 3D space.}
    \label{fig:exp_setup}
\end{figure*}

The wide availability of datasets for short-horizon manipulation tasks fostered the development of effective learning-based approaches for such tasks~\cite{levine2018learning, kalashnikov2018scalable}. In contrast, the robotics literature lacks real-world datasets for long-horizon tasks. With \textit{LHManip}, our objective is to address this existing gap in the robotics literature, providing data to develop novel approaches to solve real-world long-horizon manipulation tasks, benchmark existing methods for such tasks evaluated only in simulation~\cite{9366328} on real data, and evaluate generalization properties of state-of-the-art approaches~\cite{chen2023playfusion}. Our dataset consists of $20$ tabletop manipulation tasks involving $33$ everyday objects. For each of these tasks, we provide a natural language description and $10$ different demonstrations collected via teleoperation. Different demonstrations of the same task either involve manipulation of different object instances (e.g. objects with different texture or size) or consider different environment conditions (e.g. different distractors on the table or different initial configurations of the objects involved in the tasks). For each demonstration, we provide visual \textit{RGB} and \textit{depth} observations from a wrist-mounted and two static cameras, and robot proprioceptive information. Furthermore, for each timestep of the episode, we provide the cartesian displacement of the end-effector of the robot and the position offset applied to the gripper. Fig.~\ref{fig:lhmanip_im} shows a visual observation in \textit{LHManip} from one of the two external static cameras. This dataset forms part of the larger effort by the Open X-Embodiment collaboration project \cite{open_x_embodiment_rt_x_2023}, and this dataset paper serves to provide full details of our contribution.

\begin{figure*}[t]
    \centering
    \includegraphics[width=\linewidth]{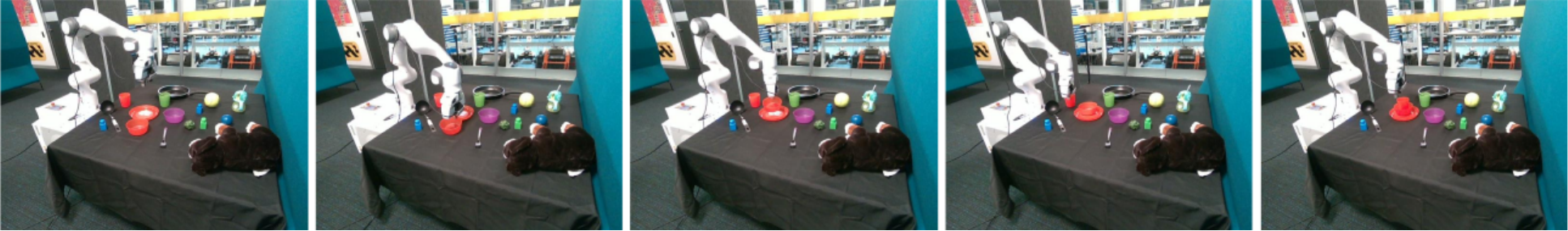}
    \caption{\textbf{Sub-tasks decomposition} of a \textit{Place the bowl on the plate and the cup in the bowl matching the color} sequence.}
    \label{fig:subtasks}
\end{figure*}

\begin{figure*}
  \centering

  \begin{minipage}{0.48\textwidth}
    \centering
    \includegraphics[width=\linewidth]{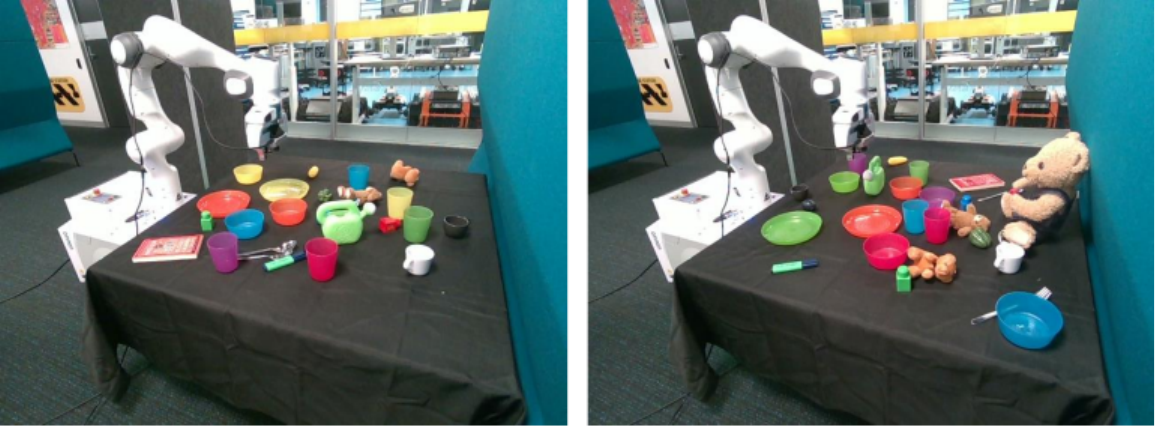}
  \end{minipage}\hfill
  \begin{minipage}{0.48\textwidth}
    \centering
    \includegraphics[width=\linewidth]{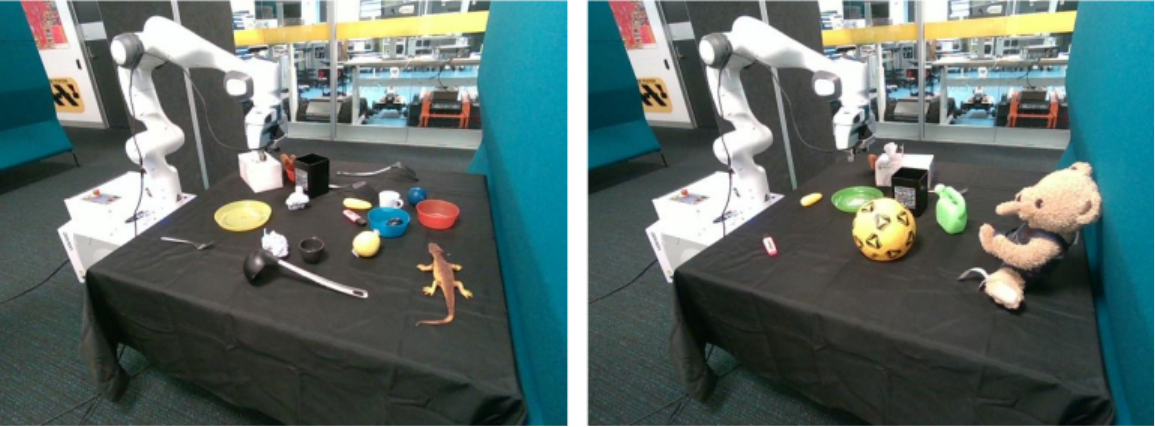}
  \end{minipage}
  \caption{\textbf{Tasks variations}: we consider different plate-bowl colors for the \textit{Place the bowls on the appropriate plates} task (left) and different plates for the \textit{Dry the plate} task (right).}
  \label{fig:variations}
\end{figure*}

\section{Related Work}
Long-horizon manipulation tasks are challenging robotics problems because they require both high-level reasoning capabilities to decompose the tasks in sequences of sub-tasks and low-level reasoning to solve the short-horizon subtasks. Existing approaches based on \textit{Hierarchical Reinforcement Learning}~\cite{NEURIPS2018_e6384711, 9366328} have been extensively studied to solve long-horizon tasks, but their application in real robotic tasks is hampered by the need of huge amounts of real training data. Other approaches solve long-horizon tasks with methods based on \textit{Imitation Learning}~\cite{Mandlekar-RSS-20}, combinations of \textit{Task and Motion Planning} (\textit{TAMP}) and \textit{Reinforcement Learning} (\textit{RL})~\cite{10227514}, or skills learning with \textit{Large Language Models} (\textit{LLM})~\cite{zhang2023bootstrap}, and would benefit from the availability of real-world data for training.

\begin{table*}[t]
\centering
\renewcommand{\arraystretch}{1.5}
\newcolumntype{C}{>{\centering\arraybackslash}m{0.3\linewidth}}
\newcolumntype{Y}{>{\centering\arraybackslash}m{0.6\linewidth}}
\begin{tabular}{C|Y}
\toprule
\textbf{\textbf{Task}} & \textbf{Success Condition} \\
\midrule
\textbf{Clean the pan.} & The robot picks the sponge and performs a movement on the pan. \\ \hline
\textbf{Cook the capsicum and place it on a plate.} & The robot picks the capsicum, puts it in the pan, and then places it on a plate. \\ \hline
\textbf{Cook the vegetables.} & All the vegetables are in the pan. \\ \hline
\textbf{Dry the plate.} & The robot grasps a tissue and performs a movement on the plate. \\ \hline
\textbf{Hide the teddy bear in the red bowl.} & The teddy bear is in the red bowl, and there is an object on it. \\ \hline
\textbf{Match the cups with the appropriate bowls.} & The cups are in the bowls of the same color. \\ \hline
\textbf{Place the bowl on the plate and the cup in the bowl matching the color.} & The bowl is on the plate of the same color, and the cup is in the bowl of the same color. \\ \hline
\textbf{Place the bowls on the appropriate plates.} & The bowls are on the plates of the same color. \\ \hline
\textbf{Prepare two cups of tea.} & The two tea bags are placed in two different cups. \\ \hline
\textbf{Put a highlighter on each book.} & There is a highlighter on each book. \\ \hline
\textbf{Put the ball in the red pot.} & The ball is in the red pot. \\ \hline
\textbf{Roll the dices in the bowl.} & The dices are thrown in the bowl. \\ \hline
\textbf{Serve the vegetables in different plates.} & The vegetables in the pan are moved to different plates. \\ \hline
\textbf{Set the table.} & The fork and the spoon are to the right of the plate, and the cup is beyond the plate from the robot's point of view. \\ \hline
\textbf{Sort the balls from left to right in order of size.} & The balls are sorted in decreasing order from left to right in the robot's point of view. \\ \hline
\textbf{Stack green blocks.} & The three green blocks are stacked. \\ \hline
\textbf{Stack the bowls.} & The three bowls on the table are stacked. \\ \hline
\textbf{Stack the cups.} & The three cups on the table are stacked. \\ \hline
\textbf{Throw away the rubbish paper.} & The rubbish paper is in the trash bin. \\ \hline
\textbf{Water the potted plant and put the can on the plate.} & The robot performs a movement with the can close to the potted plant and then puts the can on the plate. \\
\bottomrule
\end{tabular}
\caption{\textbf{Tasks}. We provide the list of tasks in the dataset, specifying the conditions that must be met to achieve them.}
\label{tab:tasks}
\end{table*}

While it has been shown in the literature the benefit of using real robotic datasets to learn short-horizon manipulation tasks such as grasping~\cite{levine2018learning}, pick and place~\cite{brohan2022rt}, pouring and scooping~\cite{zhou2023train}, object pushing~\cite{chi2023diffusionpolicy}, or combinations of different tasks~\cite{walke2023bridgedata}, datasets for long-horizon tasks are mainly provided in simulated environments. For example, the IKEA furniture assembly environment~\cite{9560986} provides simulated environments for assembly tasks that require long-horizon manipulation skills. CALVIN~\cite{9788026}, instead, provides a dataset and a benchmark for language-guided long-horizon tasks with the aim of evaluating robot capabilities of learning new skills. This differs from the ALFRED~\cite{Shridhar_2020_CVPR} benchmark, that combines seven predefined skills described through language instructions for navigation and manipulation tasks. LoHoRavens~\cite{zhang2023lohoravens}, instead, is a simulated benchmark composed of ten long-horizon language-conditioned tasks that require at least five pick-and-place steps to be achieved.

To overcome the limitations of learning robotic tasks in simulated environments, FurnitureBench~\cite{heo2023furniturebench} proposes a benchmark for long-horizon assembly tasks, such as lamp screwing and assembling table legs. While providing a huge amount of real teleoperated data ($5100$ demonstrations), the considered environments are constrained to tabletop settings with 3D-printed objects with multiple markers attached on them. The dataset used to benchmark results in~\cite{rosete2022tacorl}, instead, composes short-horizon language-guided tasks demonstrations into long-horizon tasks in constrained tabletop environments. With our dataset, we aim at overcoming these limitations and to provide a language-grounded dataset to perform cluttered tabletop manipulation tasks on everyday objects, relying only on a single language description for task description, multiple RGB-D observations of the environment and the proprioceptive state of the robot.

\section{LHManip}
\begin{table}[h]
    \vspace{.7em}
    \centering
    \begin{tabular}{ccc}
        \toprule
         & \textbf{\textit{LHManip} Items} & \\
        \midrule
        Ball & Block & Book \\
        Bowl & Capsicum & Cob \\
        Cup & Dice & Eggplant \\
        Fennel & Fork & Highlighter \\
        Ladle & Lizard & Marker \\
        Mug & Pan & Plate \\
        Plug Adapter & Plush Dog & Pot \\
        Potted Plant & Rubbish Paper & Sippy Cup \\
        Spatula & Sponge & Spoon \\
        Tea Bag & Teddy Bear & Tissue Box \\
        Trash Bin & Watering Can & Zucchini \\
        \bottomrule
    \end{tabular}
    \caption{\textbf{Items} considered in the \textit{LHManip} dataset.}
    \label{tab:objetcs}
\end{table}

\begin{table*}[t]
    \centering
    \begin{tabular}{m{0.15\linewidth}|m{0.35\linewidth}|m{0.4\linewidth}}
        \hline
        \multicolumn{1}{m{0.15\linewidth}|}{\centering \textbf{Observation}} & \multicolumn{1}{c|}{\textbf{Description}} & \multicolumn{1}{c}{\textbf{Details}} \\
        \hline
         & \centering Main static \textit{RGB} camera & \centering $640 \times 480 \times 3$ \arraybackslash \\ \cline{2-3}
         & \centering Main static \textit{Depth} camera & \centering $848 \times 480 \times 1$ \arraybackslash \\ \cline{2-3}
        \centering Cameras & \centering Secondary static \textit{RGB} camera & \centering $640 \times 480 \times 3$ \arraybackslash \\ \cline{2-3}
         & \centering Secondary static \textit{Depth} camera & \centering $848 \times 480 \times 1$ \arraybackslash \\ \cline{2-3}
         & \centering Wrist-mounted \textit{RGB} camera & \centering $640 \times 480 \times 3$ \arraybackslash \\ \cline{2-3}
         & \centering Wrist-mounted \textit{Depth} camera & \centering $848 \times 480 \times 1$ \arraybackslash \\ \cline{2-3}
        \hline
        & \centering End-effector position & \centering (x, y, z) w.r.t. root\_frame \arraybackslash \\ \cline{2-3}
        & \centering End-effector orientation & \centering (x, y, z, w) quaternion w.r.t. root\_frame \arraybackslash \\ \cline{2-3}
        \centering Proprioceptive & \centering Robot joint angles & \centering $7$ values in \textit{rad} \arraybackslash \\ \cline{2-3}
        & \centering Gripper position & \centering $2$ values in [0, 0.0404] \arraybackslash \\ \cline{2-3}
        & \centering Robot joint velocities & \centering $7$ values in \textit{rad/s} \arraybackslash \\ \hline
        \centering Instruction & \centering Natural language instruction & \centering \textit{String} \arraybackslash \\ \cline{2-3}
        \hline
        \multicolumn{1}{m{0.15\linewidth}|}{\centering \textbf{Action}} & \multicolumn{1}{c|}{\textbf{Description}} & \multicolumn{1}{c}{\textbf{Details}} \\
        \hline
        & \centering End-effector position displacement & \centering (x, y, z) w.r.t. root\_frame \arraybackslash \\ \cline{2-3}
        \centering Robot Action & \centering End-effector orientation displacement & \centering (x, y, z, w) quaternion w.r.t. root\_frame \arraybackslash \\ \cline{2-3}
        & \centering Gripper opening displacement & \centering $1$ value in [-0.0808, 0.0808] \arraybackslash \\
        \hline
    \end{tabular}
    \caption{\textbf{Observations and Actions} provided in \textit{LHManip}.}
    \label{tab:obs_act}
\end{table*}

\subsection{Experimental Set-Up and Data Collection}
We collect data via teleoperation, tracking the movements of a human operator via $10$ OptiTrack \textit{Motion Capture} (\textit{MoCap}) cameras. Fig.~\ref{fig:exp_setup} (a) shows the set-up. We equip the operator with three different sets of markers to track their movement in the cartesian space of the wrist, used to move the robot end-effector, and to measure the distance between the human thumb and index fingers, used to measure the robot gripper aperture. Fig.~\ref{fig:exp_setup} (b) shows the sets of markers used to capture the motion of the operator.

We perform the tasks on a Franka Panda 7-DoF arm~\cite{9721535} mounted on a LD-60 Omron mobile base\footnote{\url{https://www.ia.omron.com/products/family/3664/dimension.html}}, keeping the mobile base fixed throughout data collection.

We teleoperate the end-effector of the Panda arm relying on the \textit{servo} implementation of the \textit{armer}\footnote{\url{https://github.com/qcr/armer}} library. Specifically, we move the robot to mimic the human movement in the \textit{x}, \textit{y}, \textit{z} position coordinates and the \textit{yaw} orientation of the end-effector, i.e. the rotation around the axis perpendicular to the floor. Furthermore, we perform position control of the gripper joints, to mimic the fingers aperture performed by the operator.

We perform all the tasks in the dataset in the cluttered tabletop setting shown in Figs.~\ref{fig:lhmanip_im}, \ref{fig:subtasks}, and \ref{fig:variations}. We record proprioceptive information from the robot, and acquire visual and depth information from the environment. Specifically, we acquire RGB and depth information from an \textit{Intel(R) RealSense D435} mounted on the wrist of the robot and from two external \textit{Intel(R) RealSense D455}. We refer the reader to the \textit{{Observation and Action Space}} section for a detailed description of the information provided with the dataset.

\subsection{Dataset}

The proposed dataset consists of $20$ long-horizon tabletop manipulation tasks. In the following we describe the tasks and the information that we make available to the users.

\subsubsection{Tasks}

We report in Tab.~\ref{tab:tasks} an overview of the $20$ tasks considered in our dataset. The tasks are performed in a highly-cluttered tabletop environment and require the robot to manipulate everyday objects. We provide the list of objects used in the dataset in Tab.~\ref{tab:objetcs}. The considered tasks are composed by at least two sub-tasks that the robot must complete to successfully achieve the task. We report in Fig.~\ref{fig:subtasks} an exemplar sequence from the \textit{Place the bowl on the plate and the cup in the bowl matching the color} task. As it can be noticed, to achieve the task, the robot needs to perform several sub-tasks: it firstly grasps the orange bowl and places it on the orange plate. Then it picks the orange cup and places it in the orange bowl. Furthermore, with our dataset we aim at providing data to solve tasks that require high-level reasoning capabilities. In that we specify the task at hand via a natural language instruction, without providing any other low-level environment descriptions (i.e. object details like shape, spatial information, or color). We believe that the robot must be able to infer this information autonomously from the environment. For these reasons, for each task, we provide different variations of either the initial placement of the objects in the environment or variations of the objects involved to achieve the task. Fig.~\ref{fig:variations} shows two exemplar task variations for two different tasks, where the robot is required to solve the same task with different objects.

\subsubsection{Observation and Action Space}
\label{sec:obs_act_space}

\textit{LHManip} provides a set of visual and proprioceptive information that captures the robot movement and the action that the robot is required to perform via teleoperation. In the dataset, we provide for each timestep observations and actions as reported in Tab.~\ref{tab:obs_act}. We control the robot in a non-blocking mode at $30$ \textit{Hz}.

\noindent While we control the cartesian end-effector of the robot and the aperture of the gripper, in the dataset, we also provide joint-level information such as joint positions and velocities which can be used for \textit{keypoints} extraction and enable task training with keypoint-based methods as PerAct~\cite{shridhar2023perceiver}.

\noindent Please note also that, while we provide the full quaternion displacements at each timestep, we control only the rotation of the end-effector around the axis perpendicular to the floor (z quaternion value). Values different from zero in the other coordinates, therefore, aim at correct the error between the real orientation of the end-effector measured via \textit{Forward Kinematics} and the desired null orientation around the x and y axes.

\subsubsection{Dataset Access}

The \textit{LHManip} dataset is publicly available\footnote{\url{https://huggingface.co/datasets/fedeceola/LHManip}}. %and can be downloaded as a single \textit{.zip} file. %\footnote{\url{https://github.com/fedeceola/LHManip}}. % and can be downloaded as a single \textit{.zip} file. %\footnote{\url{https://www.dropbox.com/scl/fi/6t717h5mo5kyhq521qavb/long_horizon_manipulation_dataset.zip?rlkey=rk4wsxp464x5bt4a8tgz563ne&dl=0}} and can be downloaded as a single \textit{.zip} file.
We provide data in \textit{.png} format for \textit{RGB} and \textit{Depth} images, and in \textit{.pkl} files for numerical and textual information. We provide a \textit{Python} snippet code and the instructions to parse the dataset on \textit{GitHub}\footnote{\url{https://github.com/fedeceola/LHManip}}.
Furthermore, \textit{LHManip} is included as part of the larger dataset released in the \textit{Open X-Embodiment} project~\cite{open_x_embodiment_rt_x_2023}. We release additional depth and unprocessed sensory data together with this white paper, as well as the code to preprocess this dataset\footnote{\url{https://github.com/fedeceola/rlds_dataset_builder}} 
back into the desired \textit{RLDS}\footnote{\url{https://github.com/google-research/rlds}} data format as required by the \textit{Open X-Embodiment} project.

\section{Conclusion}
The resolution of long-horizon tasks is crucial for integrating robots performing everyday tasks in our homes. Motivated by the success of learning-based approaches from short-term manipulation datasets, we presented \textit{LHManip}, a dataset for long-horizon robotic manipulation tasks, with the aim of addressing the current gap in the literature where such datasets are lacking.

In the perspective of developing robots that can interact with humans in everyday environments, we believe that they must possess the ability to address long-horizon tasks based on a single high-level natural instruction, relying solely on visual and proprioceptive feedback. Existing real datasets are either constrained to simplified environments or have long-horizon instructions composed of short-horizon task descriptions. \textit{LHManip}, instead, presents several challenges, such as natural language instruction understanding, visual perception of the environment in the presence of changing challenging conditions, and learning of low-level control policies or \textit{keypoint}-based methods for sub-tasks execution.

We hope that our work will motivate the need for more datasets, benchmarks and methodologies to learn long-horizon manipulation tasks, thereby taking a significant stride toward the integration of robots into human-centered environments.

\section*{Acknowledgments}
F.C. and L.N. acknowledge financial support from the PNRR MUR project PE0000013-FAIR.
N.S. and K.R. also acknowledge the ongoing support from the QUT Centre for Robotics. N.S. and K.R. were partially supported by the Australian Government through the Australian Research Council's Discovery Projects funding scheme (Project DP220102398) and by an Amazon Research Award to N.S..

%\end{comment}

%% Use plainnat to work nicely with natbib. 

\bibliographystyle{plainnat}
\bibliography{bibliography}

\end{document}